\documentclass[11pt]{article}
\usepackage{microtype}
\usepackage{graphicx}
\usepackage{booktabs}
\usepackage{subcaption}

\usepackage
{caption}
\newif\ifarxiv
\arxivtrue %

\newif\ifperfect
\perfectfalse %

\ifarxiv
    \usepackage{EMNLP2023}
\else 
    \usepackage[review]{EMNLP2023}
\fi

\usepackage{hyperref}

\usepackage[T1]{fontenc}
\usepackage{times}
\usepackage{latexsym}
\usepackage{sec/package_zhijing}

\usepackage[utf8]{inputenc}

\usepackage{microtype}

\usepackage{inconsolata}
\usepackage{amsmath}
\usepackage{xcolor}
\usepackage{tikz}
\usepackage{pdfrender}
\usepackage{amssymb}
\usepackage{bm}
\definecolor{yellow}{HTML}{FECA57}
\definecolor{orange}{HTML}{FF9F43}
\definecolor{pink}{HTML}{FF9FF3}

\newcommand{\ccirc}[1]{\tikz\draw[red, fill=#1, draw=black, line width=.9pt] (0,0) circle (.7ex);}

\newcommand{\ourmodel}{{Moûsai}\xspace}

\newcommand{\ourdata}{\textsc{Text2Music}\xspace}

\newcommand{\oururl}{\ifarxiv \url{http://bit.ly/44ozWDH} \else \url{https://bit.ly/anonymous-mousai} \fi\xspace}

\newcommand{\ourtitle}{\textit{\ourmodel}: Efficient Text-to-Music Diffusion Models 
\xspace}
\renewcommand{\v}[1]{\pmb{#1}}
\newcommand{\defeq}{:=
}

\title{\ourtitle}

\author{%
  Flavio Schneider\thanks{{ } Co-first author. 
  \quad \quad $^\dag$Co-supervision.}
  \\
   ETH Zürich \\
  \texttt{ \href{mailto:flavio.schneider.97@gmail.com}{flavio.schneider.97@gmail.com}} 
  \And
  Ojasv Kamal\samethanks{} 
  \\
  IIT Kharagpur \\
  \texttt{ \href{mailto:kamalojasv2000@gmail.com}{kamalojasv2000@gmail.com}} \\
  \AND
  Zhijing Jin$^\dag$ 
  \\
  MPI for Intelligent Systems \& ETH Zürich \\
  \texttt{\href{mailto:jinzhi@ethz.ch}{jinzhi@ethz.ch}} \\
  \And
  Bernhard Schölkopf$^\dag$
  \\
  MPI for Intelligent Systems \\
  \texttt{\href{mailto:bs@tue.mpg.de}{bs@tue.mpg.de}} \\
}

\begin{document}
\maketitle
\begin{abstract}
Recent years have seen the rapid development of large generative models for text; however, much less research has explored the connection between text
and another ``language'' of communication -- \textit{music}. 
\ifarxiv
Music, much like text, can convey emotions, stories,
and ideas, and has its own unique structure and
syntax.
\fi
In our work, we bridge text and music via a text-to-music generation model that is highly efficient, expressive, 
and can handle long-term structure. 
Specifically, we develop \textit{\ourmodel}, 
a cascading two-stage latent diffusion model that can generate multiple minutes of high-quality stereo music at 48kHz from textual descriptions. 
Moreover, our model features high efficiency,
which enables
real-time inference on a single consumer GPU with a reasonable speed.
Through experiments and property analyses, we show our model's competence over a variety of criteria compared with existing music generation models.\ifarxiv
~Lastly, to promote the open-source culture,
we provide a collection of open-source libraries with the hope of facilitating future work in the field.\fi \footnote{
\ifarxiv
We open-source the following:

-- Codes: {\tiny \href{https://github.com/archinetai/audio-diffusion-pytorch}{\url{https://github.com/archinetai/audio-diffusion-pytorch}}}

-- Music samples for this paper:
{\tiny \oururl}

-- Music samples for all models: {\tiny \href{https://bit.ly/audio-diffusion}{\url{https://bit.ly/audio-diffusion}}}
\fi
}
\end{abstract}

\section{Introduction}

\label{introduction}

In recent years, natural language processing (NLP) has made significant strides in understanding and generating human language, due to the advancements in deep learning and large-scale pre-trained models \cite{radford2018improving,devlin-etal-2019-bert,gpt3}. While the majority of NLP research has focused on textual data, there exists another rich and expressive ``language'' of communication -- \textit{music}. Music, much like text, can convey emotions \cite{music_emotion}, stories \cite{chung2006digital}, and ideas \cite{bicknell2002can}, and has its own unique structure and syntax \cite{swain1995concept}. 

In this paper, we further bridge the gap between text and music by leveraging the power of NLP techniques to generate music conditioned on textual input. Through our work, we not only aim to expand the scope of NLP applications, but also contribute to the interdisciplinary research at the intersection of language, music, and machine learning techniques.

\begin{figure}[t]
\vskip 0.2in
\begin{center}
\centerline{\includegraphics[width=\columnwidth]{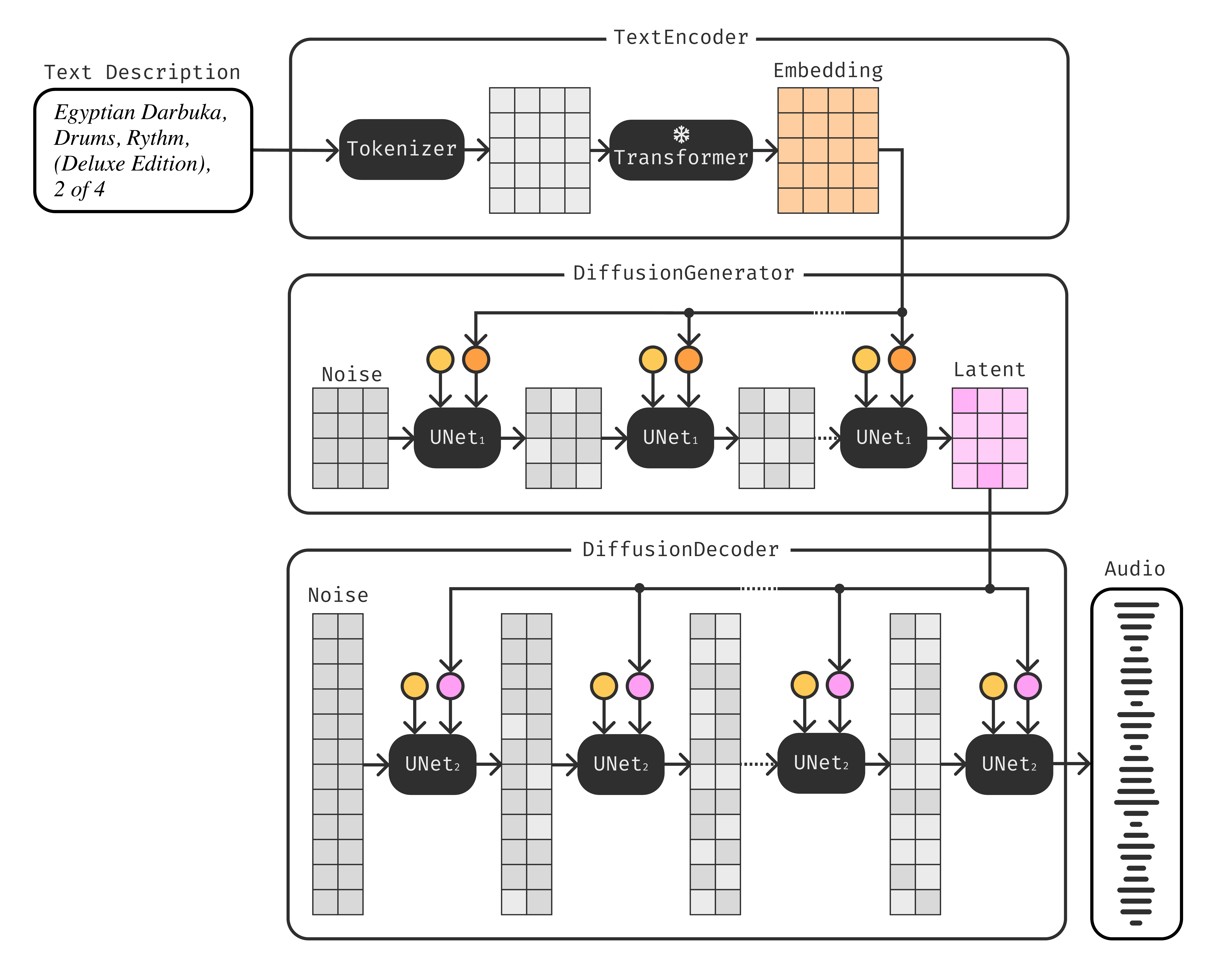}}
\caption{
We propose a two-stage cascading diffusion method, where the first stage compresses the music using a novel diffusion autoencoder, and the second stage generates music from the reduced representation conditioned on the encoding of a textual description. 
}
\label{fig1}
\end{center}
\end{figure}

However, like text, music generation has long been a challenging task, as it requires multiple aspects at different levels of abstraction \cite{WaveNet,dieleman2018challenge}. 
Existing audio generation models explore the use of recursive neural networks \cite{mehri2017samplernn}, adversarial generative networks \cite{kumar2019melgan,kim2021lip,engel2019gansynth,morrison2022chunked}, autoencoders \cite{deng2021unsupervised}, and transformers \cite{yu2022museformer}.
With the recent advancement in diffusion-based generative models in computer vision \cite{Dalle2, Imagen}, researchers in speech have also started to explore the use of diffusion models in tasks such as 
speech synthesis \cite{kong2021diffwave,lam2022bddm,leng2022binauralgrad}, although only a few these models can apply well
to the task of music generation.

Additionally, there are several long-standing challenges in the area of music generation:
(1) music generation at length, as most
text-to-audio systems \cite{Riffusion, AudioGen} can only generate \textit{a few seconds} of audio; (2) model efficiency, as many need to run on GPUs for hours to generate just one minute of audio \cite{Jukebox,AudioGen}; (3) lack of diversity of the generated music, as many are limited by their training methods taking in a single modality (resulting in the ability to handle only single-genre music, but \textit{not diverse} genres) \cite{RAVE, Musika}; and (4) easy controllability by text prompts, as most are only controlled by latent states \cite{RAVE,Musika}, the starting snippet of the music \cite{AudioLM}, or text but are lyrics \cite{Jukebox} or descriptions of a daily sound like dog barking \cite{AudioGen}.

\ifarxiv
A single model mastering all these aspects would make a strong contribution to
the music industry, as it can enable the broader public to be part of the creative process by allowing them to compose music using an accessible text-based interface, assist creators in finding inspiration, and provide an unlimited supply of novel audio samples.
\fi

To address these challenges, we propose \textit{\ourmodel},\footnote{\ourmodel is romanized ancient Greek for {\em Muses}, the sources of artistic inspiration (\url{https://en.wikipedia.org/wiki/Muses}), and also evokes a blend of {\em music} and {\em AI}.
} a novel text-conditional two-stage cascading diffusion model. Specifically, the first stage trains a music encoder by diffusion magnitude-autoencoding (DMAE), which compress audio by the novel diffusion autoencoder; and the second stage learns to generate the reduced representation while conditioning on a textual description by text-conditioned latent diffusion (TCLD).
The two-stage generation process is shown in \cref{fig1}.

Apart from proposing the novel text-to-music diffusion model, we also introduce some special designs to boost model efficiency, making the model more accessible. First, our DMAE can achieve an audio signal compression rate of 64x. Moreover, we design a lightweight and specialized 1D U-Net architecture. Together, our model achieves a fast inference speed on a single consumer GPU in minutes, and a training time of approximately one week per stage on one A100 GPU, making it possible to train and run the overall system using resources available in most universities.

We train our model on a newly collected dataset, \ourdata, with 50K text-music pairs covering diverse music genres\ifarxiv
. Remarkably, our diffusion-based model improves significantly on previous models, as it can be trained on a \textit{variety} of music genres, generate \textit{long-context} music for several minutes with a \textit{high quality of 48kHz} stereo music, run \textit{real-time} inference efficiently within minutes, and can be easily controlled by text. Our extensive evaluations on 11 criteria also validate the quality of the generated music by our model from multiple perspectives, such as text-music relevance, and music quality.
\fi

In summary, our contributions are as follows:
\begin{enumerate}[nolistsep]
    \item We are the first to propose the text-to-music diffusion model using a two-stage cascading latent diffusion modeling process.
    \item We achieve high efficiency with a compression rate of 64x, and a specialized U-Net design, which achieves a training time of one week on an A100 consumer GPU, and real-time inference time.
    \item We collect \ourdata, a dataset of 50K text-music pairs constituting 2,500 hours of music.
    \item Our model outperforms existing baselines by clear margins on 11 different evaluation criteria, demonstrating merits such as high efficiency, text-music relevance, music quality, and long-context structure.
\end{enumerate}

\section{Related Work}
\label{related-work}

\myparagraph{Connecting Text and Music}
The connection between text and music lies in the intersection of NLP and computational musicology. 
Previous work looks into aspects such as the similarity of music and linguistic structures
\cite{papadimitriou-jurafsky-2020-learning}, music and dialog \cite{berlingerio-bonin-2018-towards}, and
jointly modeling music and text for emotion detection \cite{mihalcea-strapparava-2012-lyrics}. Apart from several work that generates music from text \cite{Jukebox,Riffusion},
we are the first to explore diffusion models to interact text with music representations.

\myparagraph{Generative Models}
Generative models aim to learn a lower-dimension representation space, and then reconstruct to the high-dimension space conditioning on the given information \cite{LatentDiffusion, DiffSound, AudioGen, ImagenVideo}.
Some effective methods earlier include
auto-encoding \cite{AE, VAE}, or quantized auto-encoding \cite{VQVAE, VQGAN, RQVAE}.
Recent proposals focus on 
the quantized representation followed by masked or autoregressive learning on tokens \ifarxiv\cite{Phenaki, Parti, Muse,Jukebox, AudioLM, DiffSound, AudioGen}, 
\fi
and 
diffusion models
\ifarxiv
\cite{Dalle2, LatentDiffusion, Imagen, ImagenVideo, Riffusion}, 
\fi
which leads to impressive performance. 
To the best of our knowledge, we are the first to adapt the cascading diffusion approach for audio generation.

\section{\ourmodel:
Efficient Long-Context Music Generation from Text
}
\label{model}

Our model \ourmodel contains a two-stage training process.
In Stage 1, we use
diffusion magnitude-autoencoding (DMAE), which
compresses
the audio waveform 64x using a diffusion autoencoder. 
In 
Stage 2, we use a
latent text-to-audio diffusion model, 
to
generate a novel latent space by diffusion while conditioning on text embeddings obtained from a
frozen 
transformer language model.

\subsection{Stage 1: 
Music Encoding
by Diffusion Magnitude-Autoencoding (DMAE)}\label{sec:stage1}

We design the first step of \ourmodel to be learning a good music encoder to capture the latent representation space for music.
Representation learning is crucial for generative models, as it can be drastically more efficient than handling the high-dimensional raw input data \cite{LatentDiffusion, DiffSound, AudioGen, ImagenVideo, Phenaki}.

\begin{figure}[t]
\begin{center}
\centerline{\includegraphics[width=0.8\columnwidth]{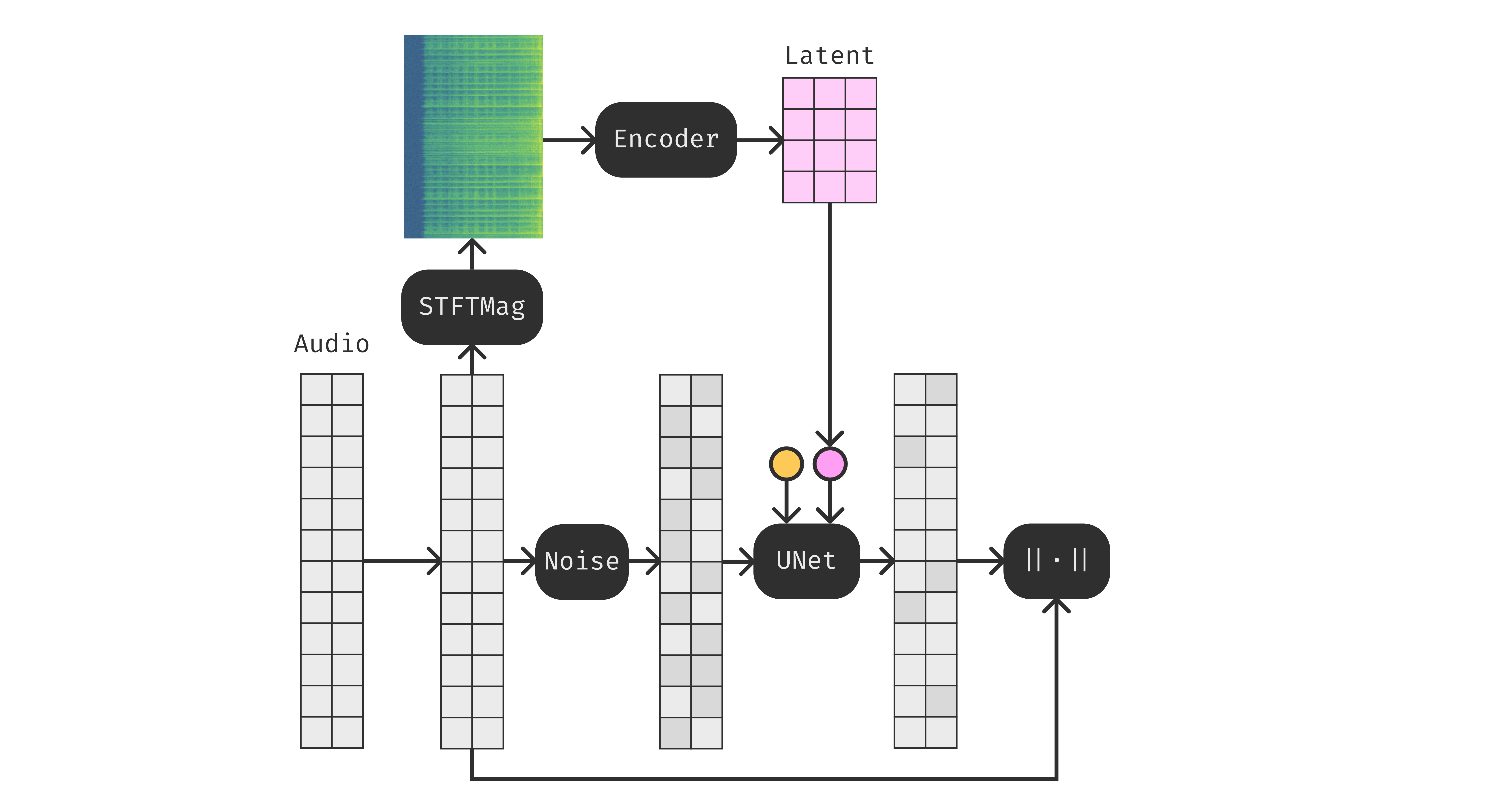}}
\caption{The training scheme of our diffusion magnitude autoencoder (DMAE).
When denoising (bottom right), we condition the U-Net on the noise level (\ccirc{yellow}) and compressed latent representation (\ccirc{pink}) from a reduced version of the non-noisy audio (the pink matrix).}
\label{fig:dmae}
\end{center}
\end{figure}

\myparagraph{Overview}
To learn the representation space for music, we deploy a diffusion magnitude autoencoder (DMAE) shown in \cref{fig:dmae}.
Specifically, 
we adopt our diffusion-based audio autoencoder, introduced in \cref{sec:audio_diff}, to compress audio into a smaller latent space by 64x from the original waveform. To train the model, we first convert the waveform to a magnitude spectrogram, which is a better representation for audio models, and then we auto-encode it into a latent representation. 

At the same time, we corrupt the original audio with a random amount of noise, and train our 1D U-Net (introduced in \cref{sec:unet}) to remove that noise. During the noise removal process, we condition the U-Net on the noise level and the compressed latent, which can have access to a reduced version of the non-noisy audio.

\subsubsection{$\v{v}$-Objective Diffusion}\label{sec:diffusion}

We use the $\v{v}$-objective diffusion process as proposed by \citet{VDiffusion}. Suppose we have a sample $\v{x}_0$ from a distribution $p(\v{x}_0)$, some noise schedule $\sigma_t\in[0,1]$, and some noisy data point $\v{x}_{\sigma_t} = \alpha_{\sigma_t}  \v{x}_0 + \beta_{\sigma_t}  \v{\epsilon}$.
The $\v{v}$-objective diffusion tries to estimate a model $\hat{\v{v}}_{\sigma_t}=f(\v{x}_{\sigma_t}, \sigma_t)$ by minimizing the following objective:
\begin{align}
    \mathbb{E}_{t\sim[0,1],\sigma_t,\mathbf{x}_{\sigma_t}}\left[\|f_{\mathbf{\theta}}(\mathbf{x}_{\sigma_t}, \sigma_t) - \mathbf{v}_{\sigma_t}\|_2^2\right]
    ~,
\end{align}

where $\v{v}_{\sigma_t}=\frac{\partial \v{x}_{\sigma_t}}{\sigma_t}=\alpha_{\sigma_t} \v{\epsilon} - \beta_{\sigma_t}\v{x}_{0}$, for which we define $\phi_t\defeq\frac{\pi}{2}\sigma_t$, and obtain its trigonometric values 
$\alpha_{\sigma_t}\defeq\cos(\phi_t)$, 
and $\beta_{\sigma_t}\defeq\sin(\phi_t)$.

\subsubsection{DDIM Sampler for Denoising}
The denoising step uses
ODE samplers
to turn noise into a new data point by estimating the rate of change. In this work, we adopt the DDIM sampler \cite{DDIM}, which we find to work well and have a reasonable tradeoff between the number of steps and audio quality. The DDIM sampler denoises the signal by repeated application of the following:
\begin{align}
    \hat{\v{v}}_{\sigma_t} &= f_\theta(\v{x}_{\sigma_t}, \sigma_t) \\ 
    \hat{\v{x}}_0 &= \alpha_{\sigma_t} \v{x}_{\sigma_t} - \beta_{\sigma_t} \hat{\v{v}}_{\sigma_t} \\
    \hat{\v{\epsilon}}_{\sigma_t} &= \beta_{\sigma_t}\v{x}_{\sigma_{t}} + \alpha_{\sigma_t}\hat{\v{v}}_{\sigma_t} \\
    \hat{\v{x}}_{\sigma_{t-1}} &= \alpha_{\sigma_{t-1}}\hat{\v{x}}_0 + \beta_{\sigma_{t-1}}\hat{\v{\epsilon}}_t,
\end{align} 
which estimates both the initial data point and the noise at the step $\sigma_t$, for some $T$-step noise schedule $\sigma_T,\dots,\sigma_0$ as a sequence evenly spaced between 1 and 0.

\subsubsection{Diffusion Autoencoder for Audio Input}\label{sec:audio_diff}
We propose a new diffusion autoencoder that first encodes a magnitude spectrogram into a compressed representation, and later injects the latent into intermediate channels of the decoding modules.
The standard method to do diffusion, such as the image diffusion model \cite{LatentDiffusion}, is to compress the input 
into a lower-dimensional representation space and apply the diffusion process on the reduced latent space.
We further compress and enhance the representation space by diffusion-based autoencoding \cite{DiffAE}, which is first introduced in computer vision,
as a way to condition the diffusion process on a compressed latent vector of the input itself. 
Since diffusion serves as a more powerful generative decoder, and hence the input can be reduced to latent representations with higher compression ratios.

\subsubsection{Efficient and Enriched 1D U-Net}\label{sec:unet}
Another crucial module in our model is the efficient 1D U-Net that we design.
We identify that the vanilla U-Net architecture \cite{UNet}, originally introduced for medial image segmentation, has relatively limited efficiency and speed, as it uses
an hourglass convolutional-only 2D architecture with skip connections.

Hence, we propose a novel
U-Net with only 1D convolutional kernels, which is more efficient than the original 2D architecture in terms of speed, and can be successfully used both on waveforms or on spectrograms if each frequency is considered as a different channel. 

\begin{figure}[t]
\begin{center}
\centerline{\includegraphics[width=\columnwidth]{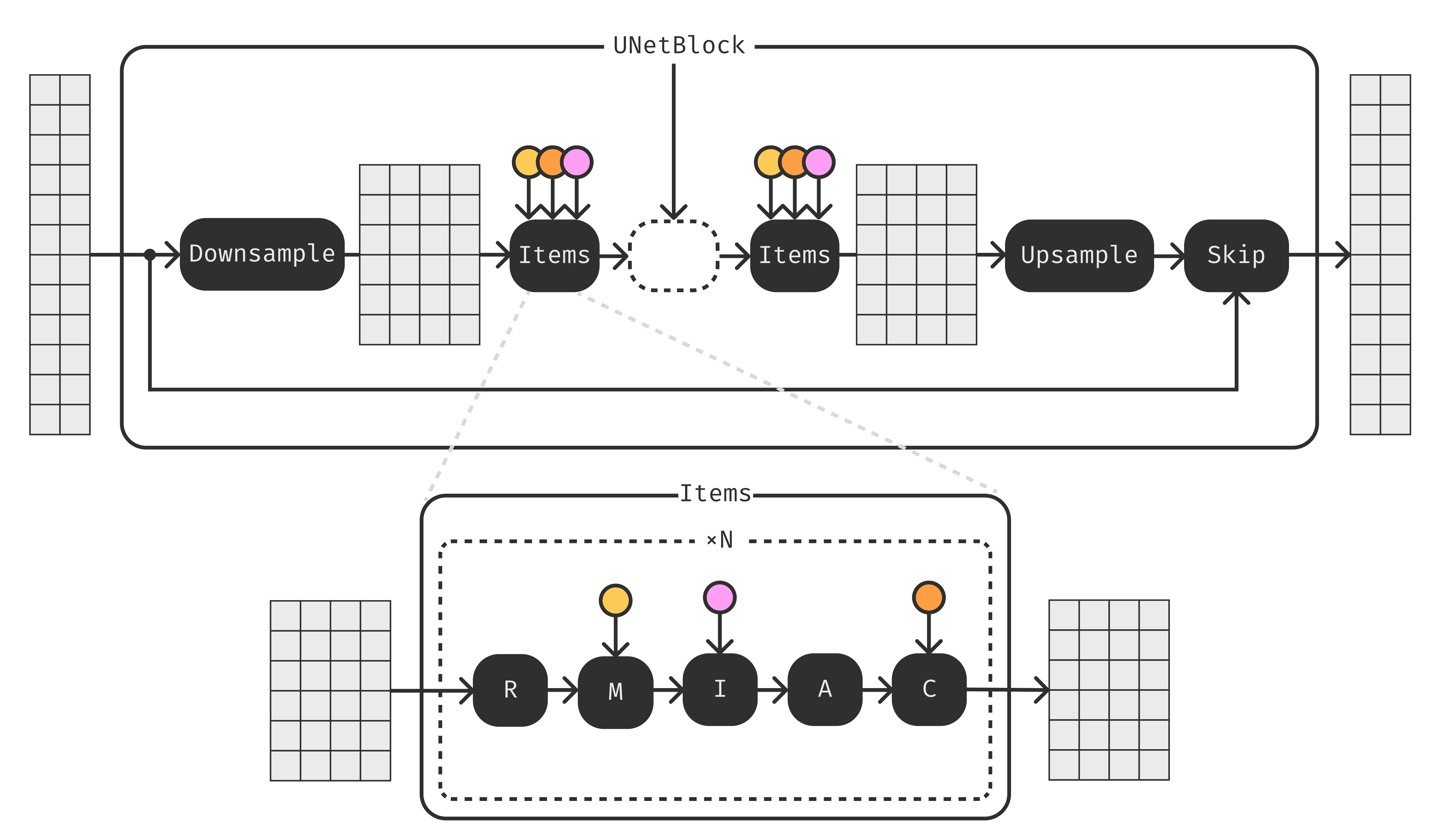}}
\caption{Our proposed 1D U-Net architecture.
Each
\texttt{UNetBlock} (top)
consists of several U-Net items (bottom).
In each U-Net item (bottom), we use a 1D convolutional ResNet (R), and a modulation unit (M)
to provide the diffusion noise level as a feature vector conditioning (\ccirc{yellow}). For Stage 1, we use an inject item (I) to inject external channels as conditioning (\ccirc{pink}), and for Stage 2, we use an attention item (A) to share time-wise information, and a cross-attention item (C) to condition on an external (text) embedding (\ccirc{orange}). 
Moreover, for the \texttt{UNetBlock}s, we can recursively nest them, which we indicate by the inner dashed region on the top.
}
\label{fig:unet-block}
\end{center}
\end{figure}

Moreover, we infuse our 1D U-Net
with multiple new components, as illustrated in \cref{fig:unet-block}:
a ResNet residual 1D convolutional unit, a modulation unit to alter the channels given features from the diffusion noise level, and an inject item to concatenate external channels to the ones at the current depth.
Note that inject items are applied only at a specific depth in the decoder in the first stage to condition on the latent representation of the music.

In summary, our novel 1D U-Net features more modern convolutional blocks, a variety of attention blocks, conditioning blocks, and improved skip connections, maintaining an efficient skeleton of the hourglass architecture.

\subsubsection{Overall Model Architecture}
Our entire Stage 1, DMAE, works as follows.
Let $\v{w}$ be a waveform of shape $[c, t]$ for $c$ channels and $t$ timesteps, and $(\v{m}_{\v{w}}, \v{p}_{\v{w}})=\operatorname{stft}(\v{w}; n=1024, h=256)$ be the magnitude and phase obtained from a short-time furier tranform of the waveform with a window size of 1024 and hop-length of 256. Then the resulting spectrograms will have shape $[c\cdot n, \frac{t}{h}]$. We discard phase and encode the magnitude into a latent $\v{z}=\mathcal{E}_{\v{\theta}_{e}}(\v{m}_{\v{w}})$ using a 1D convolutional encoder. 
The original waveform is then reconstructed by decoding the latent using a diffusion model $\hat{\v{w}}=\mathcal{D}_{\v{\theta}_{d}}(\v{z}, \v{\epsilon}, s)$, where $\mathcal{D}_{\v{\theta}_{d}}$ is the diffusion sampling process with starting noise $\v{\epsilon}$ and $s$ is the number of decoding (sampling) steps. The decoder is trained with $\v{v}$-objective diffusion while conditioning on the latent $f_{\v{\theta}_{d}}(\v{w}_{\sigma_t}; \sigma_t, \v{z})$, where $f_{\v{\theta}_{d}}$ is the proposed 1D U-Net, called repeatedly during decoding.

Since only the magnitude is used and phase is discarded, this diffusion autoencoder is simultaneously a compressing autoencoder and vocoder. By using the magnitude spectrograms, higher compression ratios can be obtained than autoencoding directly the waveform. We found that waveforms are less compressible and efficient to work with. Similarly, discarding phase is beneficial to obtaining higher compression ratios for the same level of quality. The diffusion model can easily learn to generate a waveform with realistic phase even if conditioned only on the encoded magnitude.

In this way, the latent space for music can serve as the starting point for our text-to-music generator, which will be introduced next.
To ensure this representation space fits the next stage, 
we apply a $\mathrm{tanh}$ function on the bottleneck, keeping the values in the range $[-1,1]$. 
Note that we do not use a more disentangled bottleneck, such as the one in VAEs \cite{VAE},
as its
additional regularization reduces the amount of allowed compressibility. 

\subsection{Stage 2: 
Text-to-Music Generation by Text-Conditioned Latent Diffusion (TCLD)
}\label{sec:stage2}
\begin{figure}[t]
\vskip 0.2in
\begin{center}
\centerline{\includegraphics[width=0.7\columnwidth]{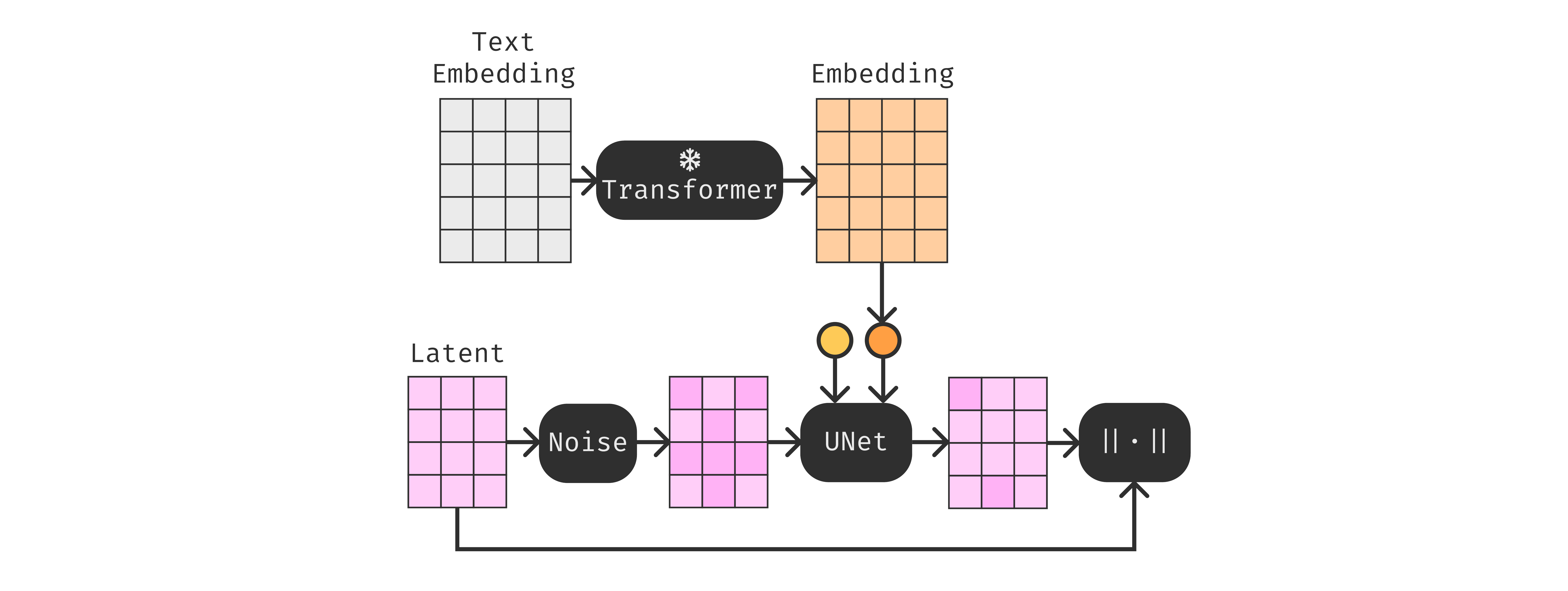}}
\caption{The training scheme of our text-conditioned latent diffusion (TCLD) generator.
During the denoising process,
we provide the U-Net
a feature vector (\ccirc{yellow}) and 
a text embedding
(\ccirc{orange}). }
\label{fig:tcld}
\end{center}
\vspace{-4mm}
\end{figure}

Based on the learned music representation space, in this stage, we guide the music generation with text descriptions.

\myparagraph{Overview}
As shown in \cref{fig:tcld},
we propose a text-conditioned latent diffusion (TCLD) process. Specifically, we first corrupt the latent space of music with a random amount of noise, then train a series of U-Nets to remove the noise, and condition the U-Nets' denoising process on a text prompt encoded by a transformer model. In this way, the generated music both conforms to the latent space of music and corresponds to the text prompt.

\subsubsection{Text Conditioning}\label{sec:text_condition}
To obtain the text embeddings, prior work on text-conditioning suggests either learning a joint data-text representation \cite{CLIP, CLAP, Dalle2}, or using embeddings from pre-trained language model as direct conditioning \cite{Imagen, ImagenVideo} of the latent model. 
In our TCLD model, 
we follow the practice in \citet{Imagen} to use a pre-trained and frozen T5 language model \cite{T5} to generate text embeddings from the given description. We use the classifier-free guidance (CFG) \cite{CFG} with a learned mask applied on batch elements with a probability of 0.1 to improve the strength of the text-embedding during inference.

\subsubsection{Adapting the U-Net for Text Conditioning}
To enable the U-Net to condition on the text embedding $\v{e}$, we append two additional blocks to the U-Net:
an attention item to share long-context structural information, and a cross-attention item to condition on the text embeddings, as in \cref{fig:unet-block}. 
These attention blocks ensure information sharing over the entire latent space, which is crucial to learn long-range audio structure.

Given the compressed size of the latent space, we also increase the size of this inner U-Net to be larger than the first stage. And due to our efficiency design, it maintains a reasonable training and inference speed, even with large parameter counts. 

\subsubsection{Overall Model Architecture}
We illustrate the detailed process in \cref{fig:tcld}.
Consistent with the previous stage,
we use $\v{v}$-objective diffusion and the 1D U-Net architecture. When condition on the text embedding $\v{e}$, 
we use the U-Net configuration $f_{\v{\theta}_{g}}(\v{z}_{\sigma_t}; \sigma_t, \v{e})$ to generate the compressed latent $\v{z}=\mathcal{E}_{\v{\theta}_{e}}(\v{m}_{\v{w}})$. 
Then, the generator $\mathcal{G}_{\v{\theta}_{g}}(\v{e}, \v{\epsilon}, s)$ applies DDIM sampling and calls the U-Net $s$ times to generate an approximate latent $\hat{\v{z}}$ from the text embedding $\v{e}$ and starting noise $\v{\epsilon}$. The final generation stack during inference to obtain a waveform is
\begin{align}
    \hat{\v{w}}=\mathcal{D}_{\v{\theta}_{d}}(\mathcal{G}_{\v{\theta}_{g}}(\v{e}, \v{\epsilon}_{g}, s_{g}), \v{\epsilon}_{d}, s_{d})
    ~.
\end{align}
\section{Experimental Setup}\label{sec:setup}
\subsection{Collection of the \ourdata Dataset}\label{sec:dataset}

To provide a fertile ground to train our text-to-music model on, we collect a new dataset, \ourdata, which consists of 50K text-music pairs totaling
2,500 hours. We ensure a high quality of stereo music sampled at 48kHz and cover a wide variety of music spanning multiple genres, artists, instruments, and provenience. Many existing open-source music datasets, such as \citet{gillick2019learning, hawthorne2018enabling}, have limitations in terms of the specific musical instruments they encompass. While some datasets, like \citet{engel2017neural, boulangerlewandowski2012modeling}, cover a broader array of instruments, they fall short in representing a wide variety of genres. This inadequacy underscores the need for a more comprehensive dataset that encompasses a rich tapestry of musical genres and diverse instrumentation.

As for the procedure to collect the music, 
\ifarxiv
we first check with the copyright regulations, which grants an exemption for using copyright infringing copies if the purpose is scientific research \cite{geiger2018exception,delacroix2023data}, according to
the EU regulation in Article 3 of the EU Directive on Copyright in the Digital Single Market \cite{proposal2016european}.
Then, 
\fi
we
follow Spotify's top recommendations to collect seven very large playlists, each containing on average
7K pieces of music.
We iterate through every music sample in these playlists, for which we use the name of the music to search and download the music from YouTube, and we use the metadata to compose its corresponding text description, which contains the music title, author, album name, genre, and year of release.

\ifarxiv
In line with our spirit to open-source the model, we also open-source the data collection pipeline on GitHub,\footnote{
\ifarxiv
\url{https://github.com/archinetai/audio-data-pytorch}
\else
Anonymous link. We will release it upon acceptance.
\fi
} so future researchers can use it to facilitate new data collection.
\fi

We show the statistics about the diverse set of genres in our \ourdata dataset in \cref{tab:data_distr}.
\begin{table}[t]
    \centering \small
    \begin{tabular}{lcccc}
    \toprule
    \textbf{Genre}    & \textbf{\# Pieces} & \textbf{Percentage (\%) in Dataset} \\ \midrule
    Pop & 5,498 & 27.29 \\
    Electronic & 3,875 & 19.38 \\
    Rock & 3,584 & 17.79 \\
    Metal & 1,796 & 8.92 \\
    Hip Hop & 818 & 4.06 \\
    Others & 4,492 & 22.56 \\
    \bottomrule
    \end{tabular}
    \caption{Our \textsc{Text2Music} dataset covers a variety of music, e.g., pop, electronic, rock, metal, hip pop, etc.}
    \label{tab:data_distr}    
\end{table}

\subsection{Implementation Details}
Our diffusion autoencoder has 185M parameters, and text-conditional generator has 857M parameters, with more architecture details in \cref{appd:impl}.
We train the music autoencoder on random crops of length $2^{18}$ ($\sim$5.5s at 48kHz), and the text-conditional diffusion generation model on fixed crops of length $2^{21}$ ($\sim$44s at 48kHz) encoded in the 32-channels, 64x compressed latent representation. 
We use the AdamW optimizer \cite{AdamW} with a learning rate of $10^{-4}$, $\beta_1$ of $0.95$, $\beta_2$ of $0.999$, $\epsilon$ of $10^{-6}$, and weight decay of $10^{-3}$. And we use an exponential moving average (EMA) with $\beta=0.995$ and power of $0.7$.

\section{Evaluation}\label{sec:results}
\begin{table*}[t]
\centering %

\setlength\tabcolsep{2.7pt}
\resizebox{\textwidth}{!}{
\begin{tabular}{lllllllllllllll}
\toprule
\textbf{Model}                                      & \textbf{Sample Rate}$\uparrow$ & \textbf{Len.}$\uparrow$ & \textbf{Input (Text\cmark)}                       & \textbf{Music (Diverse$\uparrow$)}                  & \textbf{Example}                                                  & 
{\textbf{Infer. Time$\downarrow$}}   
& \textbf{Data}
\\ \midrule
WaveNet (\citeyear{WaveNet})                                   & 16kHz@1     & Secs     & None                                  & Piano or speech                    & Piano                                                    & \textbf{= Audio len.$^\star$} 
& 260                
\\
Jukebox (\citeyear{Jukebox})                                   & 44.1kHz@1   & \textbf{Mins$^\star$}    &  \textbf{Lyrics}, author, etc.             & Song with the lyrics    & Song                                                     & Hours 
& 70K             
\\
RAVE  (\citeyear{RAVE})                                   & \textbf{48kHz@2}   & Secs$^\star$     & Latent                                  & Single-genre Music            & Strings                                                    & \textbf{= Audio len.}$^\star$
& 100         
\\
AudioLM (\citeyear{AudioLM})                                   & 16kHz@1     & Secs$^\star$    & Beginning of the music                       & Piano or speech                    & Piano                                                    & Mins 
& 40K             
\\
Musika  (\citeyear{Musika})                                   & 22.5kHz@2   & Secs     & Context vector                                  & Single-genre Music            & Piano                                                    & \textbf{= Audio len.}$^\star$
& 1K              
\\
Riffusion (\citeyear{Riffusion})                  & 44.1kHz@1   & 5s     & \textbf{Text} 
(genre, author, etc.) & 
\textbf{Music of any genre}
& Jazzy clarinet                            & Mins 
& --          
\\
AudioGen (\citeyear{AudioGen})                                  & 16kHz@1     & Secs$^\star$    & \textbf{Text} 
(a phrase/sentence)  & 
Daily sounds
& Dog barks                                           & Hours 
& 4K              
\\ \hline

\textbf{\ourmodel} (Ours) & \textbf{48kHz}@\textbf{2}     & \textbf{Mins$^\star$}    & \textbf{Text} 
(genre, author, etc.) & 
\textbf{Music of any genre}
& African drums & \textbf{= Audio len.}
& 2.5K            
\\
\bottomrule

\end{tabular}
}
\caption{Comparison of our \ourmodel model with previous music/audio generation models. We compare the followings aspects: (1) audio sample rate@the number of channels ({Sample Rate$\uparrow$}, where the higher the better), (2) context length of the generated music ({Len.$\uparrow$}, where the higher the more capable the model is to generate structural music; $^\star$ indicates variable length, where we assume that autoregressive methods are variable by default, with an upper-bound imposed by attention), (3) input type ({Input}, where we feature using \text{Text} as the condition for the generation), (4) type of the generate music ({Music}, where the more \text{Diverse$\uparrow$} genre, the better), (5) an example of the generated music type ({Example}), (6) inference time ({Infer. Time$\downarrow$}, where the shorter the better, and since the music length is seconds or minutes, the inference time equivalent to the audio length is the shortest, and we use $^\star$ to show models that can run inference fast on CPU), and (7) total length of the music in the training data in hours ({Data}).}
\label{tab:comparison}
\end{table*}
\subsection{Assessment Criteria Overview}

Evaluating music is a highly challenging task. We survey a large number of papers, and find that previous work adopts a variety of objective and subjective metrics,\footnote{The common metrics we surveyed 
include quality \cite{goel2022raw}, fidelity \cite{goel2022raw, hawthorne2019enabling, hyun2022commu}, musicality \cite{goel2022raw, yu2022museformer, Jukebox}, diversity \cite{goel2022raw, Jukebox}, and structure \cite{yu2022museformer, leng2022binauralgrad, Jukebox}.} and the gist is that {no single metric is perfect}. 
After careful thinking, we design a comprehensive set of evaluation metrics covering three categories with a total of \textit{11 metrics}, including both automatic and human evaluations.
In the following, we will introduce the overall property analysis (\cref{sec:eval_property}), such as the sample rate, prompt type, and music type; efficiency (\cref{sec:efficiency}); text-music relevance (\cref{sec:eval_relevance}); music quality (\cref{sec:eval_music}); and long-term structure of the music  (\cref{sec:prompt_analysis}).

\subsection{Property Analysis}\label{sec:eval_property}

Comparing the overall properties of various models in \cref{tab:comparison}, we see a set of impressive properties of the \ourmodel model: 
(1) We are among the very few that can control music generation easily by \textit{text descriptions} of the type of music we want, as most other models do not take text as input \cite{WaveNet,RAVE,AudioLM}, or take only 
lyrics or descriptions of daily sounds (e.g., ``a dog barking'') \cite{AudioGen,Jukebox}. The only other text-to-music model is the Riffusion model \cite{Riffusion}, which only works with very short length of 5 seconds.

(2) Our model is also among the very few that enables \textit{long-context} music generation for several minutes, among all others that can only generate seconds \cite{WaveNet,Riffusion, AudioGen,Musika}, except for Jukebox \cite{Jukebox} which generates songs given lyrics and takes very long to run inference.

(3) Moreover, we also highlight the \textit{diversity} of music we generate, as our model design enables multi-genre music training, instead of single-genre ones in previous models \cite{RAVE, Musika}, and we can find rhythm, loops, riffs, and occasionally even entire choruses in our generated music.

\subsection{Efficiency of Our Model}\label{sec:efficiency}
{Efficiency} is another highlight of our model, where we only need an inference time similar to the audio length on a consumer GPU, which is several minutes, while many other text-to-audio models take many GPU hours \cite{Jukebox,AudioGen}, as in \cref{tab:comparison}. 
Our model is very friendly for research at university labs, as 
each model can be trained on a single A100 GPU in 1 week of training using a batch size of 32.
\ifarxiv
This is equivalent to around 1M steps for both the diffusion autoencoder and latent generator. For inference, as an example, a novel audio source of {$\sim$43s} can be synthesized {in} less than {50s} using a consumer GPU with a DDIM sampler and a high step count (100 generation steps and 100 decoding steps).
\fi

We also calculate the exact inference statistics for our \ourmodel vs. Riffusion  models in \cref{tab:clap}, and find that our model needs less than 1/5 the inference time, and almost half of the inference memory than Riffusion does. 
\begin{table}[ht]
    \centering \small
\setlength\tabcolsep{2pt}
    \begin{tabular}{lcccc}
    \toprule
    \textbf{Model} & \textbf{Inf. Time (s) ($\downarrow$)} & \textbf{Mem. (G) ($\downarrow$)} & \textbf{RTF} ($\downarrow$) \\ \midrule
    Riffusion & 218.0&8.85 & 5.07 \\
    \ourmodel & \textbf{49.2} & \textbf{5.04} & 1.14 \\
    \bottomrule
    \end{tabular}
    \caption{Efficiency evaluation
of our \ourmodel and Riffusion in terms of the inference time (Inf. Time)\ifarxiv ~by seconds\fi, inference memory (Mem.) \ifarxiv by Gigabytes \fi, and the real-time factor (RTF) to generate a single 43-second music clip.
    }
    \label{tab:clap}
\end{table}

\subsection{Evaluating the Text-Music Relevance}\label{sec:eval_relevance}

To assess how much the generated music corresponds to the given text prompt, we deploy both human and automatic evaluations.

\myparagraph{Relevance \& Distinctiveness by Human Evaluation
}\label{sec:eval_diversity}
We design a listener test where the annotators need to infer some coarse information of the text prompt behind a given piece of generated music.
Since it is too challenging to infer the exact text prompt, we only ask annotators to infer the music genre indicated in the prompt.

To prepare the ground-truth prompts, we compose a list of 40 random text prompts spanning across the four most common music genres in our \ourdata dataset: electronic, hip hop, metal, and pop. See \cref{appd:prompts} for the entire list of prompts.
 ~Inspired by the two-alternative forced choice (2AFC) experiment design, we design a \textit{four-alternative forced choice (4AFC)} paradigm, where the annotators need to categorize each music sample into exactly one of the four provided categories. See annotation details in \cref{appd:annot_genre}.

\begin{figure}[t]
\centering
  \begin{subfigure}[t]{0.45\linewidth}
    \includegraphics[width=\linewidth]{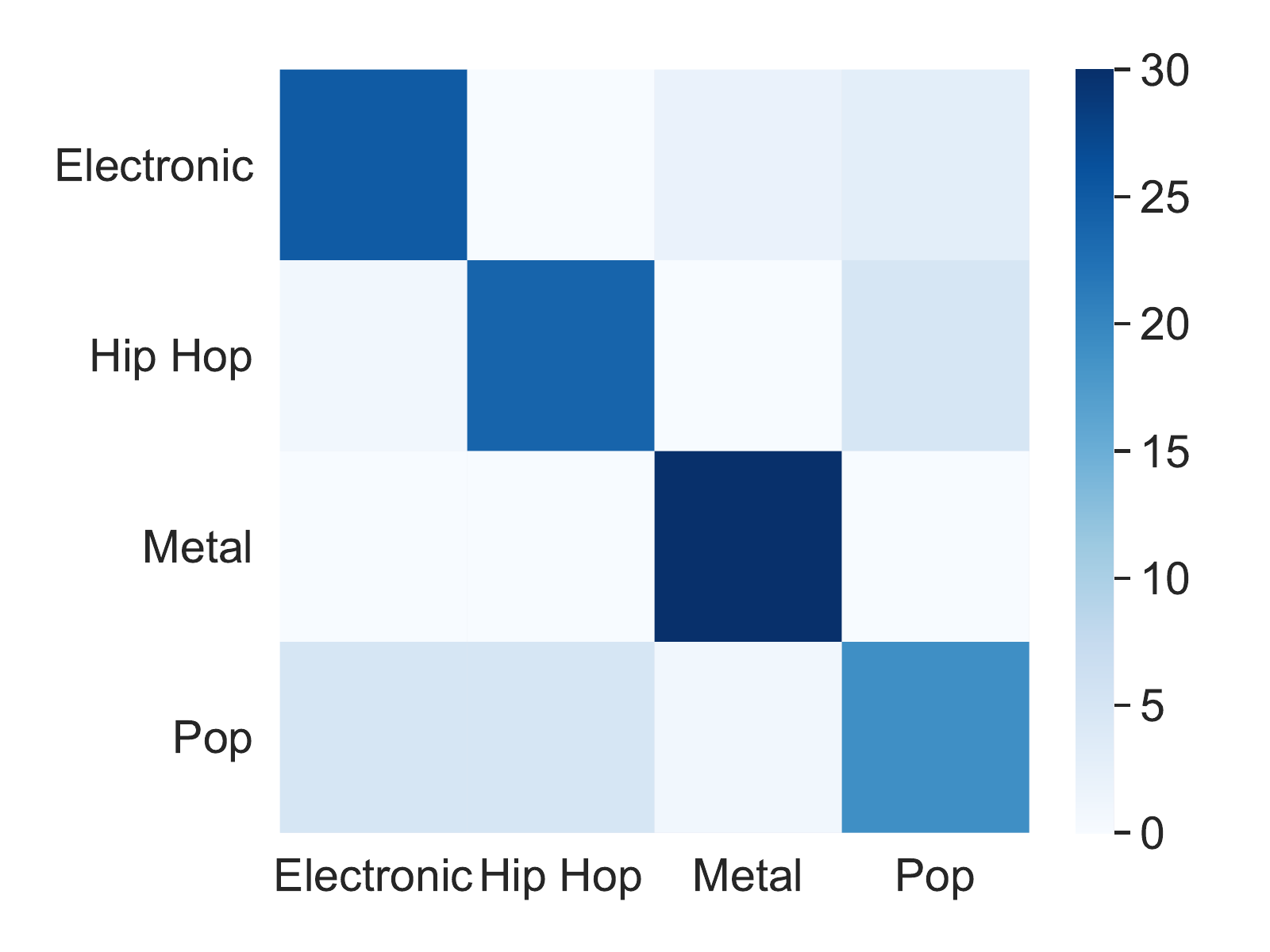}
     \caption{Confusion matrix for the music pieces generated by \ourmodel. ($y$-axis: true genre; $x$-axis: inferred genre.)}
  \end{subfigure}
    ~ 
  \begin{subfigure}[t]{0.45\linewidth}
    \includegraphics[width=\linewidth]{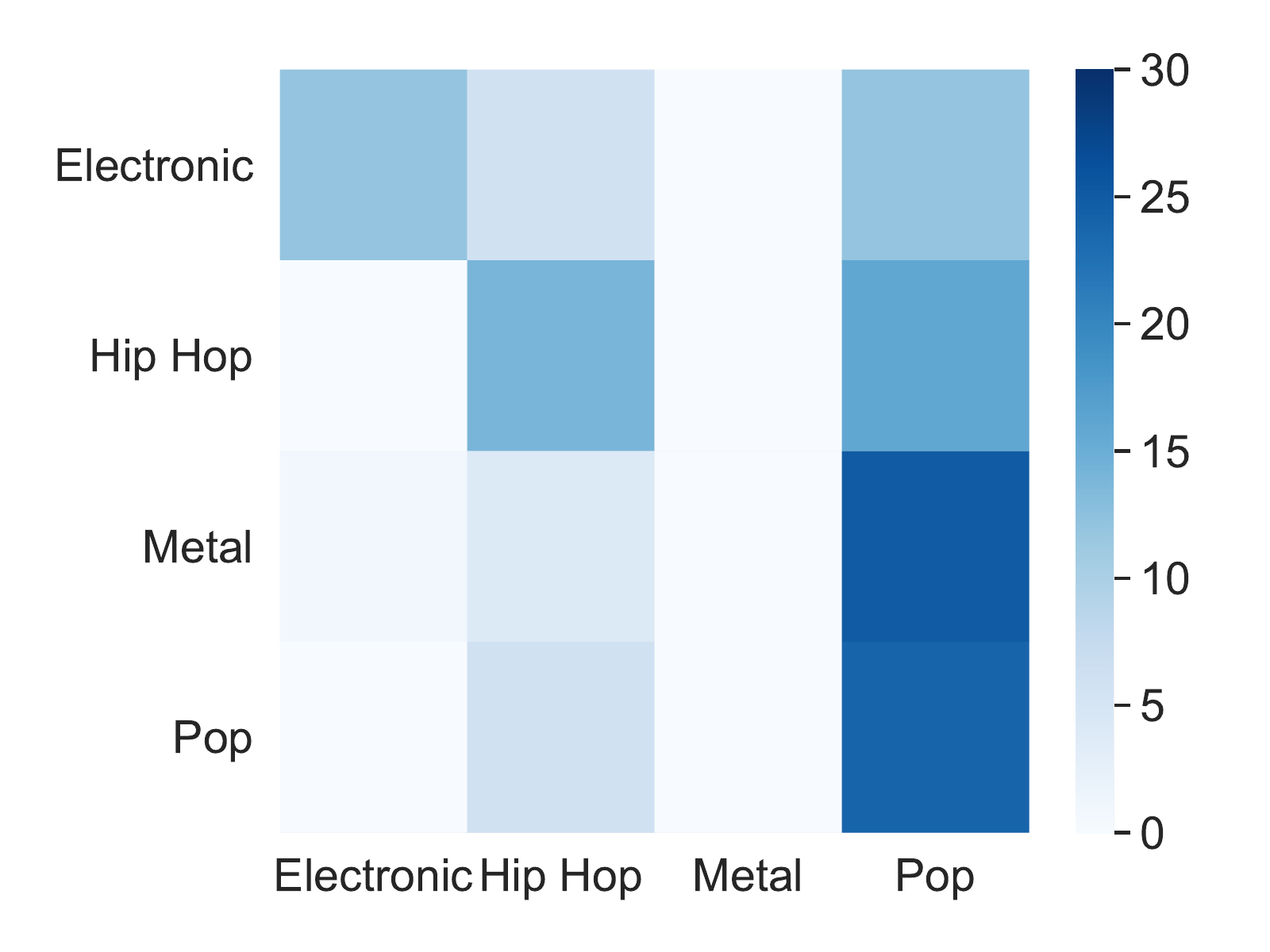}
     \caption{Confusion matrix for the music pieces generated by the Riffusion model.}
  \end{subfigure}
   \caption{For the text-music relevance check, we ask the annotators to infer the ground-truth genres of the generated music
   by
   (a) our model and (b) the Riffusion model.
The darker diagonal means better results.
   \ifarxiv
   Note that each matrix adds up to 120, corresponding to 40 samples per model annotated by three perceivers each.
   \fi
   } \label{fig:categorization}
\end{figure}

In \cref{fig:categorization}, we can see that our \ourmodel model has the most mass on the diagonal (i.e., correctly identified), while the Riffusion model tends to generate generic samples that are mostly identified as pop for all ground-truth genres.
This shows that the music generated by our model is both relevant to the test and distinct enough with the given genre against others.

\myparagraph{Relevance by CLAP}
For automatic evaluation, we adopt the commonly used CLAP score \cite{wu2023largescale}  to quantify the alignment between the generated audio and the corresponding text. 
From \cref{tab:clap}, we can see that our model is two times better than Riffusion in terms of the CLAP score, and also much faster in inference time.
\ifarxiv
Specifically, we apply the pretrained CLAP model \cite{wu2023largescale}\footnote{\url{https://github.com/LAION-AI/CLAP}} to get the embeddings of each music sample and its corresponding text prompt, and then report their cosine similarity as the CLAP score.
\fi
\begin{table}[ht]
    \centering \small
\setlength\tabcolsep{2pt}
    \begin{tabular}{lcccc}
    \toprule
    \textbf{Model}    & \textbf{CLAP Score for Text-Music Relevance ($\uparrow$)} \\ \midrule
    Riffusion & 0.06  \\
    \ourmodel  & \textbf{0.13}  \\
    \bottomrule
    \end{tabular}
    \caption{CLAP scores of our \ourmodel and Riffusion.
    }
    \label{tab:clap}
\end{table}

\subsection{Evaluating the Music Quality}
\label{sec:eval_music}

We first introduce the four evaluation metrics for music quality, and then describe the results.

\subsubsection{Metrics for Music Quality}\label{sec:metric_music}
To evaluate the quality of the generated music, we adopt four metrics: the automatic score by FAD, a music Turing test, and human evaluation on musicality and audio clarity.

For automatic evaluation, 
we deploy the widely adopted \textit{Fréchet Audio Distance (FAD)} \cite{kilgour2019frechet} to assess the fidelity of the generated music distribution in comparison to the real music distribution (i.e., how \textit{similar} the generated music is to the authentic music). To facilitate the computation of FAD, we employ the commonly used PANN model \cite{kong2020panns}\footnote{\url{https://github.com/gudgud96/frechet-audio-distance}} as a means to effectively encode the music.

Then, we also set up three human evaluations, all on a scale of 1 (the worst) to 5 (the best).
First, we let human annotators to 
assess the \textit{authenticity/fidelity} of the generated music via
a {music Turing test} \cite{goel2022raw, hawthorne2019enabling, hyun2022commu}. 
\ifarxiv
Specifically, we ask the annotators to listen to a pair of music samples at a time, and judge which one is real and which is generated. To provide a more fine-grained score, we also ask them how much the generated music they identified sounds like real music, on a scale of 1 (not similar at all) to 5 (highly similar).
We keep their annotation score if they identify the generated music correctly, and otherwise we rate the music as 5, which means that the music perfectly passes the Turing test.
\fi
See more evaluation details in \cref{appd:annot_turing}.

The other two metrics we deploy are \textit{musicality} and \textit{audio clarity}. For musicality, we let human annotators rate the melodiousness and harmoniousness \cite{seitz2005dalcroze} of the given music. And for audio clarity, or quality \cite{goel2022raw}, we let them judge how close the quality is to a walkie-talkie (worst) or a high-quality studio sound system (best). The detailed setup of all our human evaluations are in \cref{appd:annot_turing} and \cref{appd:annot_musicality}.

\subsubsection{Results}\label{sec:music_results}

We show the evaluation results on all five metrics in \cref{tab:obj_eval}. We can see that, on the automatic evaluation of FAD, our model has the best score, which is one magnitude smaller than previous models. Moreover, it also shows strong performance across the human evaluation metrics, outperforming the other two models on the music Turing test, harmoniousness, and sound clarity, as well as being comparable on the melodiousness metric. 

\begin{table}[ht]
    \centering \small
    \setlength\tabcolsep{2pt}
    \begin{tabular}{llcccccccc}
    \toprule
    \textbf{Model}    & \textbf{FAD ($\downarrow$) } & \textbf{Fidelity} & \textbf{Melody} & \textbf{Harmony} & \textbf{Clarity}
    \\ \midrule
    Riffusion 
    & 0.0018 
    & 2.8
    & 2.66 
    & 2.48
    & 2.37
    \\
    Musika 
    & 0.0020
    & 3.04
    & \textbf{3.21}
    & 3.04
    & 2.88
    \\
    \ourmodel 
    & \textbf{0.00015
    }
    & \textbf{3.17}
    & 3.15
    & \textbf{3.08}
    & \textbf{2.92}
    \\
    \bottomrule
    \end{tabular}
    \caption{
Music quality scores for
the three models.    
    }
    \label{tab:obj_eval}
\end{table}

\subsection
{Long-Term Structure of the Music}\label{sec:prompt_analysis}
\begin{figure}[h]

\centerline{\includegraphics[width=0.8\columnwidth]{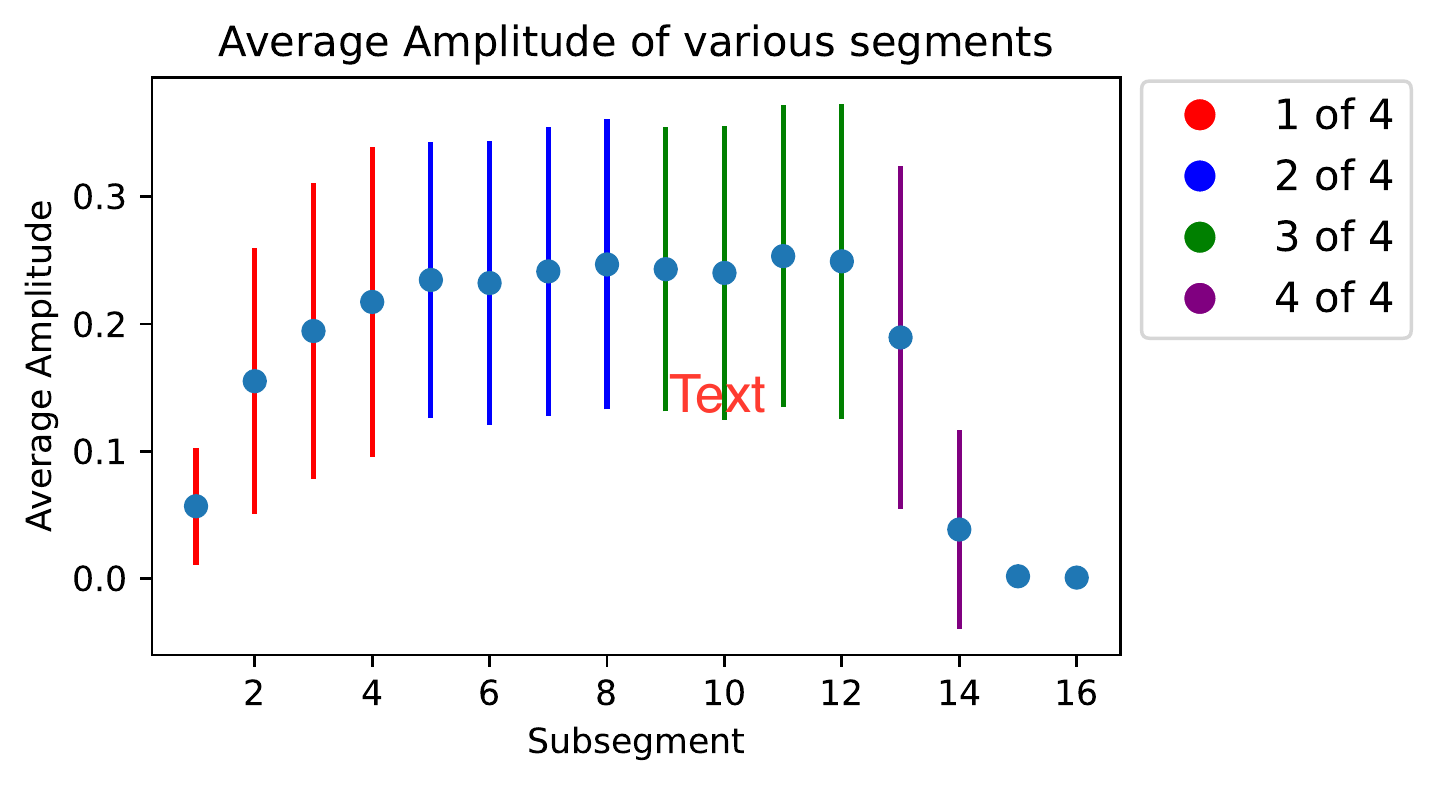}}
\caption{The average amplitude and variation of 1K random music samples spanning different segments.
}
\label{fig:avg_amp}

\end{figure}

In music composition, the arrangement of a piece typically follows a gradual introduction, a main body with the core content, and a gradual conclusion, also called the sonata form \cite{webster2001sonata}. 
Accordingly, we look into whether our generated music also shows such a long-term structure. Using the same text prompt, we can generate different segments/intervals of it by attaching the expression ``1/2/3/4 out of 4’’ at the end of the text prompt, such as ``Italian Hip Hop 2022, 3 of 4.’’
\ifarxiv We randomly generate 1,000 music pieces, where the prompts are from a uniform distribution of the four segment tags. 
\fi
We visualize the results in \cref{fig:avg_amp}, where we see the first segment shows a gradual increase in both the average amplitude and variance, followed by continuously high average amplitude and variance throughout Segments 2 and 3, and finally concluding with a gradual decline in the last segment.

\subsection{Effect of Hyperparameters}
We also explore the effect of different hyperparameters, and find that increasing the number of attention blocks (e.g., from a total of 4--8 to a total of 32+) in the latent diffusion model can improve the general structure of the songs, thanks to the long-context view. Also, if the model is trained without attention blocks, the context provided by the U-Net is not large enough to learn any meaningful long-term structure. We describe other variations of hyperparameters and findings in \cref{appd:model_variation}.

\section{Conclusion}
\label{conclusion}

In this work, we presented \ourmodel, a novel text-to-music generation model using latent diffusion.
We show that, in contrast to earlier approaches, our model can generate minutes of music in real-time on a consumer GPU, with good music quality and text-audio binding. 
\ifarxiv In addition, we provide a collection of open-source libraries to facilitate future work in the field. \fi
The work helps pave the way towards higher-quality, longer-context text-to-music generation for future applications.

\section*{Limitations and Future Work}
\label{future-work}

\myparagraph{Data Scale}
Enhancing the scale of both data and the model holds promising potential for yielding significant improvements in quality. Following \cite{Jukebox, AudioLM}, we suggest training with 50K-100K hours instead of 2.5K.
Computer Vision studies like \citet{Imagen} show that utilizing larger pretrained language models for text embeddings plays an important role in achieving better quality outcomes. Drawing upon this, we hypothesize that the application of a larger pretrained language model to our second-stage model can similarly contribute to enhanced quality outcomes.

\myparagraph{Models}
Some promising future modelling approaches that can be explored in future work include: (1) training diffusion models using perceptual losses on the waveforms instead of L2 --- this might help decrease the initial size of the U-Net, as we would not have to process non-perceivable sounds, (2) improving the quality of the diffusion autoencoder by using mel-spectrograms instead of magnitude spectrograms as input, (3) other types of conditioning which are not text-based might be useful to navigate the audio latent space, which is often hard to describe in words --- DreamBooth-like models \cite{Ruiz2022DreamBoothFT}, and (4)
more sophisticated diffusion samplers to achieve higher quality for the same number of sampling steps, or similarly more advanced distillation techniques \cite{VDiffusion}.

\ifarxiv
\section*{Author Contributions}\label{sec:contributions}
\textbf{Flavio Schneider} came up with the idea, made many architecture innovations, trained the model, and wrote a significant portion of the paper. This model is one of the several models he proposed as part of his Master's thesis at ETH Zürich \cite{schneider2023ArchiSound}.

\textbf{Ojasv Kamal}
deployed our models on the classical music data, 
set up the baseline models, %
and conducted most of the analyses for the automatic evaluation and human evaluation.

\textbf{Zhijing Jin} co-supervised this work and Flavio's Master's thesis, conducted weekly meetings, helped designed the structure of the paper, led a set of human evaluation of the models, and contributed significantly to the writing.

\textbf{Bernhard Schölkopf} supervised the work and provided precious suggestions during the design process of this work, as well as extensive suggestions on the writing.

\section*{Acknowledgment}

Our project would not have been made possible without the GPU resources from various parties: We thank Shehzaad Dhuliawala for helping us set up GPU access at ETH Zürich. We thank Vincent Berenz and Lidia Pavel for setting up our GPU access at Max Planck Institute. We thank Stability AI for their generous support for the computational resources for our first version of the model. We appreciate the writing help from Rada Mihalcea to polish this paper with the idea that text and music are both a form of language. We are also grateful for the generous help by our annotators including Yuen Chen, Andrew Lee, Aylin Gunal, Fernando Gonzalez, and Yiwen Ding. We thank Nasim Rahaman for early-stage discussions to improve the model design and contributions. We thank Fernando Gonzalez and Zhiheng Lyu for helping to improve the format of the paper. 

This material is based in part upon works supported by the German Federal Ministry of Education and Research (BMBF): Tübingen AI Center, FKZ: 01IS18039B; and by the Machine Learning Cluster of Excellence, EXC number 2064/1 – Project number 390727645.
Zhijing Jin is supported by PhD fellowships from the Future of Life Institute and Open Philanthropy, as well as the travel support from ELISE (GA no 951847) for the ELLIS program. 

\fi

\section*{Ethical Considerations}

Our work aims to bridge the gap between text and music generation, enabling the creation of expressive and high-quality music from textual descriptions. While this research has the potential to benefit various applications, such as music therapy, entertainment, and education, we recognize that it may also raise concerns in terms of copyright, cultural appropriation, and the potential misuse of generated content.

\textit{Copyright and Intellectual Property:} Our model may generate music that resembles existing copyrighted works, which could lead to potential legal disputes. First of all, for research-only use, it is exempted from copyright infringement, as we mentioned in the data collection section previously. For other purposes, we suggest incorporating mechanisms to detect and avoid generating music that closely resembles copyrighted material.

\textit{Economic Impact on Musicians and Composers:} The widespread adoption of text-to-music generation models may have economic implications for musicians and composers, potentially affecting their livelihoods. We believe that our model should be used as a tool to augment and inspire human creativity, rather than replace it. We encourage collaboration between AI researchers, musicians, and composers to explore new ways of integrating AI-generated music into the creative process, ensuring that the technology benefits all stakeholders.

In conclusion, we are committed to conducting our research responsibly and ethically. We encourage the research community to engage in open discussions about the ethical implications of text-to-music generation models and to develop guidelines and best practices for their responsible use. By addressing these concerns, we hope to contribute to the development of AI technologies that benefit society and promote creativity, while respecting the rights and values of all stakeholders.

\bibliography{custom,sec/refs_ai_safety}
\bibliographystyle{acl_natbib}

\cleardoublepage
\appendix
\section{More Data Details}
\subsection{Data Collection Rationale}
We have several desiderata when collecting the dataset. The data must (1) have text data paired with the music piece, and (2) consistitute a \textit{large} size, which means that our data crawling procedure needs to be scalable, without tedious manual efforts to curate. Note that it is crucial to get a large-sized dataset in order to unleash the performance of  audio generation diffusion models.

\subsection{Training Setup for the Text-Music Pairs} \label{training_setup}
For the textual description, we use metadata such as the title, author, album, genre, and year of release.
Given that a song could span longer than 44s, we append a string indicating which chunk is currently being trained on, together with the total chunks the song is made of (e.g., \textit{1 of 4}). This allows to select the region of interest during inference.
Hence, an example prompt is like \textit{``Egyptian Darbuka, Drums, Rythm, (Deluxe Edition), 2 of 4.''}
To make the conditioning more robust, we shuffle the list of metadata and drop each element with a probability of $0.1$. Furthermore, for 50\% of the times we concatenate the list with spaces and the other 50\% of the times we use commas to make the interface more robust during inference. Some example prompts in our dataset can be seen in \cref{tab:dataset}.

\begin{table}[h]
    \centering \small
    \begin{tabular}{p{7.5cm}ll}
\toprule
\textbf{Example Text Prompts in Our Dataset} \\ \midrule
    Nr. 415 (Premium Edition), german hip hop, 2 of 7, 2012, XATAR, Konnekt \\ \hline
    30 Años de Exitos, Mundanzas, 2 of 6, latin pop, Lupita D'Alessio, 2011 \\ \hline
    emo rap 2018 Runaway Lil Peep 4 of 5 \\ \hline
    Alone, Pt. II (Remixes) 2020 electro house Alone, Pt. II - Da Tweekaz Remix Alan Walker \\
    \bottomrule
    \end{tabular}
    \caption{Example text prompts in our dataset.}
    \label{tab:dataset}
\end{table}

\begin{table}[ht]
    \centering \small
    \begin{tabular}{p{7.4cm}lll}
    \toprule
    \textbf{\textit{Genre = Electronic}}     \\
-- Drops, Kanine Remix, Darkzy, Drops Remixes, bass house, (Deluxe) (Remix) 3 of 4 \\
-- Electronic, Dance, EDM (Deluxe) (Remix) 3 of 4 \\
-- Electro House (Remix), 2023, 3 of 4 \\
-- Electro Swing Remix 2030 (Deluxe Edition) 3 of 4 \\
-- Future Bass, EDM (Remix) 3 of 4, Remix \\
-- EDM (Deluxe) (Remix) 3 of 4 \\
-- EDM, Vocal, Relax, Remix, 2023, 8D Audio \\
-- Hardstyle, Drop, 8D, Remix, High Quality, 2 of 4 \\
-- Dubstep Insane Drop Remix (Deluxe Edition), 2 of 4 \\
-- Drop, French 79, BPM Artist, Vol. 4, Electronica, 2016 \\
\midrule
    \textbf{\textit{Genre = Hip Hop}}     \\
-- Real Hip Hop, 2012, Lil B, Gods Father, escape room, 3 of 4  \\
-- C'est toujours pour ceux qui savent, French Hip Hop, 2018 (Deluxe), 3 of 4 \\
-- Dejando Claro, Latin Hip Hop 2022 (Deluxe Edition) 3 of 4 \\
-- Latin Hip Hop 2022 (Deluxe Edition) 3 of 4 \\
-- Alternative Hip Hop Oh-My, 2016, (Deluxe), 3 of 4 \\
-- Es Geht Mir Gut, German Hip Hop, 2016, (Deluxe), 3 of 4 \\
-- Italian Hip Hop 2022 (Deluxe Edition) 3 of 4 \\
-- RUN, Alternative Hip Hop, 2016, (Deluxe), 3 of 4 \\
-- Hip Hop, Rap Battle, 2018 (High Quality) (Deluxe Edition) 3 of 4 \\
-- Hip Hop Tech, Bandlez, Hot Pursuit, brostep, 3 of 4 \\
\midrule
    \textbf{\textit{Genre = Metal}}     \\
-- Death Metal, 2012, 3 of 4 \\
-- Heavy Death Metal (Deluxe Edition), 3 of 4 \\
-- Black Alternative Metal, The Pick of Death (Deluxe), 2006, 3 of 4 \\
-- Kill For Metal, Iron Fire, To The Grave, melodic metal, 3 of 4 \\
-- Melodic Metal, Iron Dust (Deluxe), 2006, 3 of 4 \\
-- Possessed Death Metal Stones (Deluxe), 2006, 3 of 4 \\
-- Black Metal Venom, 2006, 3 of 4 \\
-- The Heavy Death Metal War (Deluxe), 2006, 3 of 4 \\
-- Heavy metal (Deluxe Edition), 3 of 4 \\
-- Viking Heavy Death Metal (Deluxe), 2006, 3 of 4 \\
\midrule
    \textbf{\textit{Genre = Pop}}     \\
-- (Everything I Do), I Do It For You, Bryan Adams, The Best Of Me, canadian pop, 3 of 4 \\
-- Payphone, Maroon 5, Overexposed, Pop, 2021, 3 of 4 \\
-- 24K Magic, Bruno Mars, 24K Magic, dance pop, 3 of 4  \\
-- Who Is It, Michael Jackson, Dangerous, Pop (Deluxe), 3 of 4 \\
-- Forget Me, Lewis Capaldi, Forget Me, Pop Pop, 2022, 3 of 4 \\
-- Pop, Speak Now, Taylor Swift, 2014, (Deluxe), 3 of 4 \\
-- Pop Pop, Maroon 5, Overexposed, 2016, 3 of 4 \\
-- Pointless, Lewis Capaldi, Pointless, Pop, 2022, 3 of 4 \\
-- Saved, Khalid, American Teen, Pop, 2022, 3 of 4 \\
-- Deja vu, Fearless, Pop, 2020, (Deluxe), 3 of 4 \\
\bottomrule
    \end{tabular}
    \caption{Text prompts composed for the four common music genres: electronic, hip hop, metal, and pop.}
    \label{tab:prompt_full_list}
\end{table}

\begin{table}[ht]
\centering
\small
\begin{tabular}{
p{7.5cm}}
\toprule
\textbf{\textit{Prompt}} \\
\midrule
-- Aerials, System Of A Down, Toxicity, 2001, 2 of 4 \\
-- Aloo Gobi, Weezer, OK Human, 2021, 1 of 4 \\
-- Bananas and Blow, Ween, White Pepper, 3 of 4 \\
-- Blue Light, Bloc Party, Silent Alarm, 2005, 1 of 4 \\
-- Break-Thru, Dirty Projectors, Lamp Lit Prose, 2018, 3 of 4 \\
-- B:/ Start Up, Blank Banshee, Blank Banshee 0, Future Funk, 4 of 4 \\
-- Carrion Crawler, Thee Oh Sees, Carrion Crawler / The Dream, 2011, 4 of 4 \\
-- Change, Tears For Fears, The Hurting, 1983, 3 of 4 \\
-- Comic, Bo Burnham, INSIDE (DELUXE), 4 of 4 \\
-- Eaten by Worms, Nothing, 2016, 2 of 4 \\
-- Effect and Cause, The White Stripes, Icky Thump, 2007, 4 of 4 \\
-- Feeling Good, Muse, 2001, 4 of 4 \\
-- Hip Hop, Tyler, The Creator, IGOR, 2019, 3 of 4 \\
-- Ice, The Microphones, It Was Hot, We Stayed in the Water, 2000, 2 of 4 \\
-- In My Pocket, Temples, 2 of 4 \\
-- Liquid State, Muse, 2012, 4 of 4 \\
-- Moonlight, Death From Above 1979, Outrage! Is Now, 2017, 2 of 4 \\
-- Mutilated Lips, Ween, The Mollusk, 1997, 2 of 4 \\
-- My Kind of Woman, Mac DeMarco, 2, 2012, 2 of 4 \\
-- Night Shop, Optiganally Yours, O.Y. in Hi-Fi, 2018, 3 of 4 \\
-- Red Eye Flashes Twice, Jeffery Dallas, 2010, 3 of 4 \\
-- Reflektor, Arcade Fire, Reflektor, 2013, 2 of 4 \\
-- Some Thing's Coming, I Monster, Neveroddoreven, 2005, 3 of 4 \\
-- Spaceship, Kanye West/GLC/Consequence, The College Dropout, 1 of 4 \\
-- Superfast Jellyfish (feat. Gruff Rhys and De La Soul), Gorillaz/Gruff Rhys/De La Soul, Plastic Beach, 2010, 2 of 4 \\
-- The Well and the Lighthouse, Arcade Fire, Neon Bible, 2007, 4 of 4 \\
-- Well, You Can Do It Without Me, Fear Fun, 4 of 4 \\
-- Upgrade (A Baymar College College), Deltron 3030/Del The Funky Homosapien/Dan The Automator/Kid Koala, Deltron 3030, 2000, 4 of 4 \\
-- Way We Won't, Grandaddy, Last Place, 2017, 3 of 4 \\
-- Safe From Heartbreak (if you never fall in love), Wolf Alice, 2 of 4 \\
-- Black Alternative Metal, The Pick of Death (Deluxe), 2006, 3 of 4 \\
-- Death Metal, 2012, 3 of 4 \\
-- Drops, Kanine Remix, Darkzy, Drops Remixes, bass house, (Deluxe) (Remix), 3 of 4 \\
-- EDM (Deluxe) (Remix), 3 of 4 \\
-- Electro House (Remix), 2023, 3 of 4 \\
-- Electro Swing Remix 2030 (Deluxe Edition), 3 of 4 \\
-- Future Bass, EDM (Remix), Remix, 3 of 4 \\
-- Hip Hop Tech, Bandlez, Hot Pursuit, brostep, 3 of 4 \\
-- Italian Hip Hop 2022 (Deluxe Edition), 3 of 4 \\
-- Heavy metal (Deluxe Edition), 3 of 4 \\
-- The Heavy Death Metal War (Deluxe), 2006, 3 of 4 \\
-- Pop, Taylor Swift, Speak Now, 2014, (Deluxe), 3 of 4 \\
-- Melodic Metal, Iron Dust (Deluxe), 2006, 3 of 4 \\
-- Electronic, Dance, EDM (Deluxe) (Remix), 3 of 4 \\
-- Alternative Hip Hop Oh-My, 2016, (Deluxe), 3 of 4 \\
-- Viking Heavy Death Metal (Deluxe), 2006, 3 of 4 \\
-- Possessed Death Metal Stones (Deluxe), 2006, 3 of 4 \\
-- Hardstyle, Drop, 8D, Remix, High Quality, 2 of 4 \\
-- Drop, French 79, BPM Artist, Vol. 4, Electronica, 2016 \\
-- Dubstep Insane Drop Remix (Deluxe Edition), 2 of 4 \\
\bottomrule
\end{tabular}
\caption{Text prompts used to generate music for the musicality test.}
\label{tab:prompts_musicality}
\end{table}

\subsection{Model Architecture and Parameters}\label{appd:impl}
Our diffusion autoencoder has 185M parameters, with 7 nested U-Net blocks of increasing channel count ([256, 512, 512, 512, 1024, 1024, 1024]), for which we downsample each time by 2, except for the first block ([1, 2, 2, 2, 2, 2, 2]). This makes the compression factor for our autoencoder to be 64x. Depending on the desired speed/quality tradeoff, more or less compression can be applied in this first stage. Following our single GPU constraint, we find that 64x compression factor is a good balance to make sure the second stage can work on a reduced representation. We discuss more about this tradeoff in \cref{comp_tradeoff}. The diffusion autoencoder only uses ResNet and modulation items with the repetitions [1, 2, 2, 2, 2, 2, 2]. We do not use attention, to allow decoding of variable and possibly very long latent representations. Channel injection only happens at depth 4, which matches the output of the magnitude encoder latent, after applying the $\mathrm{tanh}$ function.

Our text-conditional generator has 857M parameters (including the parameters of the frozen T5-base model) with 6 nested U-Net blocks of increasing channel counts ([128, 256, 512, 512, 1024, 1024]), and again downsampling each time by 2, except for the first block ([1, 2, 2, 2, 2, 2]). We use attention blocks at the depths [0, 0, 1, 1, 1, 1], skipping the first two blocks to allow for further downsampling before sharing information over the entire latent, instead use cross-attention blocks at all resolutions ([1, 1, 1, 1, 1, 1]).
For both attention and cross-attention, we use 64 head features and 12 heads per layer. We repeat items with an increasing count towards the inner U-Net low-resolution and large-context blocks ([2, 2, 2, 4, 8, 8]), this allows good structural learning over minutes of audio.

\section{More Experimental Details}
\subsection{Hardware Requirements}
\label{sec:hardware}
We use limited computational resources as available in a university lab.
Efficiency is a  highlight of our model, where we only need an inference time equivalent to the audio length on a consumer GPU, which is several minutes, while many other text-to-audio models take many GPU hours \cite{Jukebox,AudioGen}. 
Our model is very friendly for research at university labs, as 
each of our models can be trained on a single A100 GPU in 1 week of training using a batch size of 32; this is equivalent to around 1M steps for both the diffusion autoencoder and latent generator. For inference, as an example, a novel audio source of {$\sim$43s} can be synthesized {in} less than {50s} using a consumer GPU with a DDIM sampler and a high step count (100 generation steps and 100 decoding steps).

\section{More Evaluation Details}
\subsection{Annotation Details for the Genre Identification Test}
\label{appd:annot_genre}

\myparagraph{Prompts}\label{appd:prompts}
We design a listener test to illustrate the diversity and text relevance of \ourmodel. 
Specifically, we compose a list of 40 text prompts spanning across the four most common music genres in our dataset: electronic, hip hop, metal, and pop. 
The four genres are the most prevalent ones in our \ourdata dataset, which is collected from various top playlists on Spotify. Among the 50K music samples, we identify the genre of them and report the distribution of genre in \cref{tab:data_distr}. The four genres that we select for the evaluation are top in the data distribution of different genres.

When composing the test samples, we also make efforts to ensure comprehensive coverage. We first read through the text samples in the dataset and then compose samples that are reasonable music descriptions but do not exist in the data. The text prompts comprehensively cover all genres that we evaluate and incorporate essential metadata such as song title, album, artist, and year. Combining these elements allows us to generate diverse and comprehensive test prompts.
We list all the text prompts composed for the four common music genres in \cref{tab:prompt_full_list}.

Using these prompts, we generate music with both \ourmodel and the Riffusion model \cite{Riffusion}, with a total of 80 pieces of music, two for each prompt.

\myparagraph{Annotation}
To validate this quantitatively, we conducted a listener test with
three perceivers (annotators) with diverse demographic backgrounds (both female and male, 
all with at least a Bachelor's degree of education). Each annotator listens to all 80 music samples we provide, and is instructed to categorize each sample into exactly one of the four provided genres. 

We record how many times the perceiver correctly identifies the genre which the respective model was generating from.
A large number (or score) means that the model often generated music that, according to the human perceiver, plausibly belonged to the correct category (when compared to the other three categories). 
To achieve a good score, the model needs to generate diverse and genre-specific music.
We take the score as a quality score of the model when it comes to correctly performing text-conditional music generation.

\subsection{Annotation Details for Turing Test}\label{appd:annot_turing}

We conduct an evaluation employing an experiment with a similar spirit to the Turing test \cite{10.1093/mind/LIX.236.433} for natural language, but commonly called as the fidelity test in audio evaluation \cite{hyun2022commu} or speaker test \cite{greshler2021catchawaveform, hawthorne2019enabling} in audio evaluation. 

We let the annotators listen to a pair of music samples at a time, and judge which one is real and which is generated. To provide a more fine-grained score, we also ask them how much the generated music they identified sounds like real music, on a scale of 1 (almost not similar at all) to 5 (highly similar).
We keep their annotation score if they identify the generated music correctly, and otherwise we rate the music as 5, which means that the music perfectly passes the Turing test.

As for the details, we create 90 music samples, including 15 generated samples paired with 15 real music samples for each of the three models (Riffusion, Musika, and \ourmodel). We recruit two undergraduate annotators
who
have pursued playing music as a hobby for the past 10 years. 
The annotators were compensated with 500 rupees ($\sim$6.5 dollars) for this 3 hour task (which is well above daily minimum wage in India).

We presented a group of expert annotators with a total of 60 distinct folders, 15 corresponding to each of Mousai, Mousai (classical-only), Riffusion, and Musika models. Each folder contains two music files, one being the original and the other generated using a given model prompted with its corresponding metadata.

Following are the exact instructions provided to the annotators:
\begin{enumerate}
    \item You will be presented with batches of two audio samples in subfolders of this folder named from 1 to 60. Each subfolder contains two audios named a.wav and b.wav.
    \item Listen to each sample carefully.
    \item It's best to use headphones in a quiet environment if you can.
    \item Some files may be loud, so it's recommended to keep the volume moderate.
    \item One of the audio samples in each pair is a real recording, while the other is a generated (synthetic) audio.
    \item Listen to each pair of audio samples carefully.
    \item Pay attention to the quality, characteristics, and nuances of each audio sample.
    \item This folder contains a spreadsheet file called ‘Response\_Task\_2.xlsx’. Compare the samples to each other and provide a relative rating to the fake audio only out of 5, where 1 being the most fake and 5 being most real. 

\end{enumerate}

\subsection{Annotation Details for Musicality}\label{appd:annot_musicality}

In order to ascertain the quality and artistic merit of the generated musical output, we conduct a human evaluation. First, we prepare a total of 50 folders, each containing three distinct audio files, and present them to the human evaluators. We design the prompts in \cref{tab:prompts_musicality} , and run our model and all the baseline models on them.
We recruit two undergraduate annotators in India, 
who have pursued playing music as a hobby for the past 10 years. The annotators were compensated with 500 rupees ($\sim$6.5 dollars) for this 3 hour task (which is well above daily minimum wage in India).

Following are the exact instructions provided to the annotators:

\begin{enumerate}
    \item Listen to the music and rate it based on three aspects: Quality, Melody, and Harmony.
    \item It's best to use headphones in a quiet environment if you can.
    \item Some files may be loud, so it's recommended to keep the volume moderate.
    \item This folder contains folders subfolders through 1-50. Each subfolders contains three audio files named A.wav, B.wav, and C.wav. You need to listen to each of them and rate them (relative to each other) based on quality, melody, and harmony. 
    \item For Quality, consider how clear the audio sounds. Does it resemble a walkie-talkie (bad quality) or a high-quality studio sound system (good quality)?
    \item \href{https://en.wikipedia.org/wiki/Melody}{Melodiousness} refers to the main pitch or note in the music. Pay attention to the rhythm and repetitiveness of the melody. A more rhythmic and repetitive melody is considered better, while the opposite is true for a less rhythmic melody.
    \item \href{https://en.wikipedia.org/wiki/Harmony}{Harmoniousness} involves multiple notes played together to support the melody. Evaluate if these notes are in sync and enhance the effect of the melody. Higher scores should be given for good harmony and lower for poor harmony.
    \item It is recommended view youtube videos: \href{https://www.youtube.com/watch?v=xugt0hF6CNs&ab_channel=yiroubassstudio}{this} or \href{https://www.youtube.com/watch?v=kG-C_Boxjxk&pp=ygUSbWVsb2R5IGFuZCBoYXJtb255&ab_channel=TinyTero}{this} short video explaining melody and harmony
    \item This folder also contains a spreadsheet by the name “Response\_Task\_1.xlsx”. Remember to provide ratings (out of 5) for each aspect of your evaluation in the file against appropriate folder number. Feel free to listen to each sample as many times before rating them.

\end{enumerate}

\section{Exploring Variations of the Model Architecture and Training Setup}\label{appd:model_variation}

\subsection{High-Frequency Sounds}\label{sec:eval_quality}

\begin{figure}[t]
\vskip 0.1in
\begin{center}
\centerline{\includegraphics[width=\columnwidth]{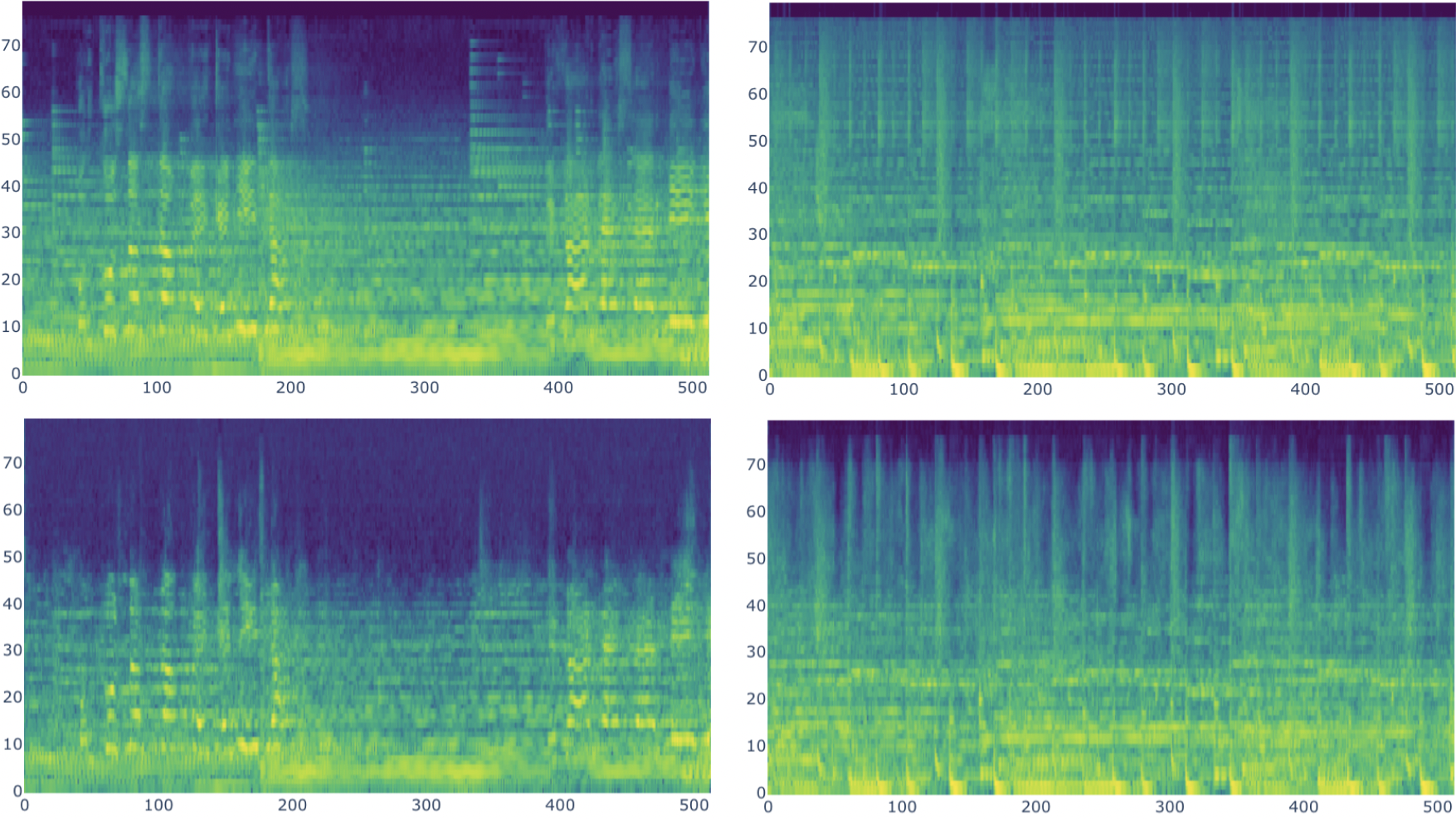}}
\caption{Mel spectrogram comparison between the true samples (top) and the auto-encoded samples (bottom); cf.\ text.
}
\label{spectrogram}
\end{center}
\end{figure}

We observe that our model is good at handling low-frequency sounds. From the mel spectrograms \cref{spectrogram}, and also the music samples we provide, we notice that our model performs well with drum-like sounds
as frequently found in electronic, house, dubstep, techno, EDM, and metal music. This is likely a consequence of the lower amount of information required to represent low-frequency sounds.

\subsection{Improving the Structure}\label{sec:eval_structure}
We find that increasing the number of attention blocks (e.g., from a total of 4 -- 8 to a total of 32+) in the latent diffusion model can improve the general structure of the songs, thanks to the long-context view. If the model is trained without attention blocks, the context provided by the U-Net is not large enough to learn any meaningful long-term structure.

\subsection{Text-Audio Binding}
We find that the text-audio binding works well with CFG higher than $3.0$. Since the model is trained with metadata such as title, album, artist, genre, year, and chunk, the best keywords to control the generation appear to be frequent descriptive names, such as the genre of the music, or descriptions commonly found in titles, such as \textit{``remix''}, \textit{``(Deluxe Edition)''}, and possibly many more. A similar behavior has been observed and exploited in text-to-image models to generate better looking results.

\subsection{Trade-Off between Speed and Quality}
We find that 10 sampling steps in both stages can be enough to generate reasonable audio. 
We can achieve improved quality and reduced noise for high-frequency sounds by trading off the speed, i.e., increasing the number of sampling steps in the diffusion decoder, e.g., 50 -- 100 steps) will similarly improve the quality, likely due to the more detailed generated latents, and at the same time result in an overall better structured music. To make sure the results are comparable when varying the number of sampling steps, we use the same starting noise in both stages.
In both cases, this suggests that using more advanced samplers could be helpful to improve on the speed-quality trade-off. 

\subsection{Trade-Off between Compression Ratio and Quality}
We find that decreasing the compression ratio of the first stage (e.g., to 32x) can improve the quality of low-frequency sounds, but in turn will slow down the model, as the second stage has to work on higher dimensional data. As proposed later in \cref{future-work}, we hypothesize that using perceptually weighted loss functions instead of L2 loss during diffusion could help this trade-off, giving a more balanced importance to high frequency sounds even at high compression ratios. \label{comp_tradeoff}

\subsection{High-Frequency Audio Generation
}\label{sec:classical_music}

We have encountered challenges in achieving satisfactory results when dealing with high-frequency audio signals, as detailed in \cref{sec:eval_quality}. To gain deeper insights into the underlying issues, we conducted an ablation experiment by exclusively training our model on classical music, a genre known for its prominent high-frequency characteristics. We train this model using 500 hours of music collected from albums of top classical composers {\cite{classical-music}} and other popular Spotify playlists. We notice a drop of 9.5\% in the fidelity score of the generated music samples compared to those produced by our original model. Further, qualitative analysis reveals that melodic elements of these samples demonstrated commendable accuracy, the harmony notes appeared to be convoluted and disorganized. This finding highlights the significance of harmonization challenges when generating high-frequency audio and underscores the need for developing improved models in future research. 

\end{document}